\documentclass[twoside,11pt]{article}
\usepackage[preprint]{jmlr2e}
\usepackage{tcolorbox}
\usepackage{color-edits}
\usepackage[scaled=0.8]{DejaVuSansMono}
\usepackage{enumitem}
\usepackage{booktabs}
\usepackage{float}
\usepackage{listings}
\usepackage{xcolor}
\ShortHeadings{LangFair}{Bouchard et al.}

\usepackage{xcolor}

\definecolor{dark-blue}{rgb}{0.15,0.15,0.4}
\definecolor{codegreen}{rgb}{0,0.6,0}
\definecolor{codegray}{rgb}{0.5,0.5,0.5}
\definecolor{codepurple}{rgb}{0.58,0,0.82}
\definecolor{backcolour}{rgb}{0.95,0.95,0.92}

\lstdefinestyle{mystyle}{
    backgroundcolor=\color{backcolour},   
    commentstyle=\color{codegreen},
    keywordstyle=\color{magenta},
    numberstyle=\tiny\color{codegray},
    stringstyle=\color{codepurple},
    basicstyle=\ttfamily\scriptsize\color{blue!30!black},    emph={int,char,double,float,unsigned,void,bool},
    emphstyle={\color{blue}},
    morekeywords={>,<,.,;,-,!,=,~},
    otherkeywords={>,<,.,;,-,!,=,~},
    breakatwhitespace=false,         
    breaklines=false,             
    language=python,
    captionpos=b,                    
    keepspaces=true,                   
    showspaces=false,                
    showstringspaces=false,
    showtabs=false,                  
    tabsize=4
}

\firstpageno{1}
\begin{document}
\title{LangFair: A Python Package for Assessing Bias and Fairness in Large Language Model Use Cases}

\author{\name Dylan Bouchard\textsuperscript{1} \email dylan.bouchard@cvshealth.com \\
        \name Mohit Singh Chauhan\textsuperscript{1} \email mohitsingh.chauhan@cvshealth.com \\
        \name David Skarbrevik\textsuperscript{1} \email david.skarbrevik@cvshealth.com \\
        \name Viren Bajaj\textsuperscript{1} \email bajajv@aetna.com \\
        \name Zeya Ahmad\textsuperscript{1} \email zeya.ahmad@cvshealth.com \\
        \addr
            The authors are the current maintainers of LangFair, and additionally have the following affiliations:
        \\
        \addr
            \textsuperscript{1}CVS Health Corporation
       }

\maketitle

\begin{abstract}
Large Language Models (LLMs) have been observed to exhibit bias in numerous ways, potentially creating or worsening outcomes for specific groups identified by protected attributes such as sex, race, sexual orientation, or age. To help address this gap, we introduce \texttt{langfair}, an open-source Python package that aims to equip LLM practitioners with the tools to evaluate bias and fairness risks relevant to their specific use cases. The package offers functionality to easily generate evaluation datasets, comprised of LLM responses to use-case-specific prompts, and subsequently calculate applicable metrics for the practitioner's use case. To guide in metric selection, LangFair offers an actionable decision framework.
\end{abstract}


\section{Introduction}
Traditional machine learning (ML) fairness toolkits like AIF360 \citep{aif360-oct-2018}, Fairlearn \citep{Weerts_Fairlearn_Assessing_and_2023}, Aequitas \citep{2018aequitas} and others \citep{vasudevan20lift, DBLP:journals/corr/abs-1907-04135, tensorflow-no-date} have laid crucial groundwork. These toolkits offer various metrics and algorithms that focus on assessing and mitigating bias and fairness through different stages of the ML lifecycle. While the fairness assessments offered by these toolkits include a wide variety of generic fairness metrics, which can also apply to certain LLM use cases, they are not tailored to the generative and context-dependent nature of LLMs.\footnote{The toolkits mentioned here offer fairness metrics for classification. In a similar vein, the recommendation fairness metrics offered in FaiRLLM \citep{Zhang_2023} can be applied to ML recommendation systems as well as LLM recommendation use cases.}

LLMs are used in systems that solve tasks such as recommendation, classification, text generation, and summarization. In practice, these systems try to restrict the responses of the LLM to the task at hand, often by including task-specific instructions in system or user prompts. When the LLM is evaluated without taking the set of task-specific prompts into account, the evaluation metrics are not representative of the system's true performance. Representing the system's actual performance is especially important when evaluating its outputs for bias and fairness risks because they pose real harm to the user and, by way of repercussions, the system developer.

Most evaluation tools, including those that assess bias and fairness risk, evaluate LLMs at the model-level by calculating metrics based on the responses of the LLMs to static benchmark datasets of prompts \citep{rudinger-EtAl:2018:N18, zhao-2018, webster-etal-2018-mind, levy2021collecting, nadeem2020stereoset, bartl2020unmasking, nangia2020crows, felkner2024winoqueercommunityintheloopbenchmarkantilgbtq, barikeri2021redditbiasrealworldresourcebias, kiritchenko2018examininggenderracebias, qian2022perturbationaugmentationfairernlp, Gehman2020RealToxicityPromptsEN, bold_2021, huang2023trustgptbenchmarktrustworthyresponsible, nozza-etal-2021-honest, parrish-etal-2022-bbq, li-etal-2020-unqovering, 10.1145/3576840.3578295}  that do not consider prompt-specific risks and are often independent of the task at hand. Holistic Evaluation of Language Models (HELM) \citep{liang2023holisticevaluationlanguagemodels}, DecodingTrust \citep{wang2023decodingtrust}, and several other toolkits \citep{srivastava2022beyond, huang2024trustllm, eval-harness, Arshaan_Nazir_and_Thadaka_Kalyan_Chakravarthy_and_David_Amore_Cecchini_and_Thadaka_Kalyan_Chakravarthy_and_Rakshit_Khajuria_and_Prikshit_Sharma_and_Ali_Tarik_Mirik_and_Veysel_Kocaman_and_David_Talby_LangTest_A_comprehensive_2024, huggingface-no-date} follow this paradigm. 

In this paper, we introduce \texttt{langfair}, an open-source Python package for conducting bias and fairness assessments of LLM use cases.\footnote{The repository for \texttt{langfair} can be found at \url{https://github.com/cvs-health/langfair}.} LangFair complements the aforementioned frameworks because it follows a bring your own prompts (BYOP) approach, which allows users to tailor the bias and fairness evaluation to their use case by computing metrics using LLM responses to user-provided prompts. This addresses the need for a task-based bias and fairness evaluation tool that accounts for prompt-specific risk for LLMs.\footnote{Experiments in \citet{wang2023decodingtrust} demonstrate that prompt content has substantial influence on the likelihood of biased LLM responses.} To guide the selection of bias and fairness metrics, LangFair offers an actionable decision framework, discussed in detail in the project's companion paper, \citet{bouchard2024actionableframeworkassessingbias}.

Furthermore, LangFair is designed for real-world LLM-based systems that require governance audits. LangFair focuses on calculating metrics from LLM responses only, which is more practical for real-world testing where access to internal states of model to retrieve embeddings or token probabilities is difficult. An added benefit is that output-based metrics, which are focused on the downstream task, have shown to be potentially more reliable than metrics derived from embeddings or token probabilities \citep{goldfarbtarrant2021intrinsicbiasmetricscorrelate, delobelle-etal-2022-measuring}.

\section{Generation of Evaluation Datasets}
The \texttt{langfair.generator} module offers two classes, \texttt{ResponseGenerator} and \texttt{CounterfactualGenerator}, which aim to enable user-friendly construction of evaluation datasets for text generation use cases.

\paragraph{\texttt{ResponseGenerator} class.}
To streamline generation of evaluation datasets, the \texttt{ResponseGenerator} class wraps an instance of a \texttt{langchain} LLM and leverages asynchronous generation with \texttt{asyncio}. To implement, users simply pass a list of prompts (strings) to the \texttt{ResponseGenerator.generate\_responses} method, which returns a dictionary containing prompts, responses, and applicable metadata.

\paragraph{\texttt{CounterfactualGenerator} class.}
In the context of LLMs, counterfactual fairness can be assessed by constructing counterfactual input pairs \citep{gallegos2024biasfairnesslargelanguage, bouchard2024actionableframeworkassessingbias}, comprised of prompt pairs that mention different protected attribute groups but are otherwise identical, and measuring the differences in the corresponding generated output pairs. These assessments are applicable to use cases that do not satisfy fairness through unawareness (FTU), meaning prompts contain mentions of protected attribute groups. To address this, the \texttt{CounterfactualGenerator} class offers functionality to check for FTU, construct counterfactual input pairs, and generate corresponding pairs of responses asynchronously using a \texttt{langchain} LLM instance.\footnote{In practice, a FTU check consists of parsing use case prompts for mentions of protected attribute groups.} Off the shelf, the FTU check and creation of counterfactual input pairs can be done for gender and race/ethnicity, but users may also provide a custom mapping of protected attribute words to enable this functionality for other attributes as well.

\section{Bias and Fairness Evaluations for Focused Use Cases}
Following \citet{bouchard2024actionableframeworkassessingbias}, evaluation metrics are categorized according to the risks they assess (toxicity, stereotypes, counterfactual unfairness, and allocational harms), as well as the use case task (text generation, classification, and recommendation).\footnote{Note that text generation encompasses all use cases for which output is text, but does not belong to a predefined set of elements (as with classification and recommendation).} Table 1 maps the classes contained in the \texttt{langfair.metrics} module to these risks. These classes are discussed in detail below.

\begin{table}[H]
\centering
\caption{Classes for Computing Evaluation Metrics in \texttt{langfair.metrics}}
\label{tab:metrics}
\begin{tabular}{lll}
Class                                           & Risk Assessed             & Applicable Tasks           \\
\toprule
\texttt{ToxicityMetrics}       & Toxicity                  & Text generation \\
\texttt{StereotypeMetrics}     & Stereotypes               & Text generation \\
\texttt{CounterfactualMetrics} & Counterfactual fairness & Text generation \\
\texttt{RecommendationMetrics} & Counterfactual fairness & Recommendation  \\
\texttt{ClassificationMetrics} & Allocational harms        & Classification  \\     
\bottomrule
\end{tabular}
\end{table}

\paragraph{Toxicity Metrics}
The \texttt{ToxicityMetrics} class facilitates simple computation of toxicity metrics from a user-provided list of LLM responses. These metrics leverage a pre-trained toxicity classifier that maps a text input to a toxicity score ranging from 0 to 1 \citep{Gehman2020RealToxicityPromptsEN, liang2023holisticevaluationlanguagemodels}. For off-the-shelf toxicity classifiers, the \texttt{ToxicityMetrics} class provides four options: two classifiers from the \texttt{detoxify} package, \texttt{roberta-hate-speech-dynabench-r4-target} from the \texttt{evaluate} package, and \texttt{toxigen} available on HuggingFace.\footnote{\url{https://github.com/unitaryai/detoxify}; \url{https://github.com/huggingface/evaluate}; \url{https://github.com/microsoft/TOXIGEN}} For additional flexibility, users can specify an ensemble of the off-the-shelf classifiers offered or provide a custom toxicity classifier object.

\paragraph{Stereotype Metrics}
To measure stereotypes in LLM responses, the \texttt{StereotypeMetrics} class offers two categories of metrics: metrics based on word cooccurrences and metrics that leverage a pre-trained stereotype classifier. Metrics based on word cooccurrences aim to assess relative cooccurrence of stereotypical words with certain protected attribute words. On the other hand, stereotype-classifier-based metrics leverage the \texttt{wu981526092/Sentence-Level-Stereotype-Detector} classifier available on HuggingFace \citep{zekun2023auditinglargelanguagemodels} and compute analogs of the aforementioned toxicity-classifier-based metrics \citep{bouchard2024actionableframeworkassessingbias}.\footnote{\url{https://huggingface.co/wu981526092/Sentence-Level-Stereotype-Detector}}

\paragraph{Counterfactual Fairness Metrics for Text Generation}
The \texttt{CounterfactualMetrics} class offers two groups of metrics to assess counterfactual fairness in text generation use cases. The first group of metrics leverage a pre-trained sentiment classifier to measure sentiment disparities in counterfactually generated outputs (see \citet{huang2020reducingsentimentbiaslanguage} for further details). This class uses the \texttt{vaderSentiment} classifier by default but also gives users the option to provide a custom sentiment classifier object.\footnote{\url{https://github.com/cjhutto/vaderSentiment}} The second group of metrics addresses a stricter desiderata and measures overall similarity in counterfactually generated outputs using well-established text similarity metrics \citep{bouchard2024actionableframeworkassessingbias}.

\paragraph{Counterfactual Fairness Metrics for Recommendation}
The \texttt{RecommendationMetrics} class is designed to assess counterfactual fairness for recommendation use cases. Specifically, these metrics measure similarity in generated lists of recommendations from counterfactual input pairs. Metrics may be computed pairwise \citep{bouchard2024actionableframeworkassessingbias}, or attribute-wise \citep{Zhang_2023}.

\paragraph{Fairness Metrics for Classification}
When LLMs are used to solve classification problems, traditional machine learning fairness metrics may be applied, provided that inputs can be mapped to a protected attribute. To this end, the \texttt{ClassificationMetrics} class offers a suite of metrics to address unfair classification by measuring disparities in predicted prevalence, false negatives, or false positives. When computing metrics using the \texttt{ClassificationMetrics} class, the user may specify whether to compute these metrics as pairwise differences \citep{aif360-oct-2018} or pairwise ratios \citep{2018aequitas}.

\section{Semi-Automated Evaluation}
\paragraph{\texttt{AutoEval} class.} 
To streamline assessments for text generation use cases, the \texttt{AutoEval} class conducts a multi-step process (each step is described in detail above) for a comprehensive fairness assessment. Specifically, these steps include metric selection (based on whether FTU is satsified), evaluation dataset generation from user-provided prompts with a user-provided LLM, and computation of applicable fairness metrics. To implement, the user is required to supply a list of prompts and an instance of \texttt{langchain} LLM. Below we provide a basic example demonstrating the execution of \texttt{AutoEval.evaluate} with a \texttt{gemini-pro} instance.\footnote{Note that this example assumes the user has already set up their VertexAI credentials and sampled a list of prompts from their use case prompts.}

\begin{lstlisting}[language=Python, caption=\texttt{AutoEval} Example]
from langchain_google_vertexai import ChatVertexAI
from langfair.auto import AutoEval

llm = ChatVertexAI(model_name='gemini-pro')
auto_object = AutoEval(prompts=prompts, langchain_llm=llm)
results = await auto_object.evaluate()
\end{lstlisting}

Under the hood, the \texttt{AutoEval.evaluate} method 1) checks for FTU, 2) generates responses and counterfactual responses (if FTU is not satisfied), and 3) calculates applicable metrics for the use case.\footnote{The `AutoEval` class is designed specifically for text generation use cases. Applicable metrics include toxicity metrics, stereotype metrics, and, if FTU is not satisfied, counterfactual fairness metrics.} This process flow is depicted in Figure 1.

\begin{figure}[H]
    \centering
    \includegraphics[width=\linewidth]{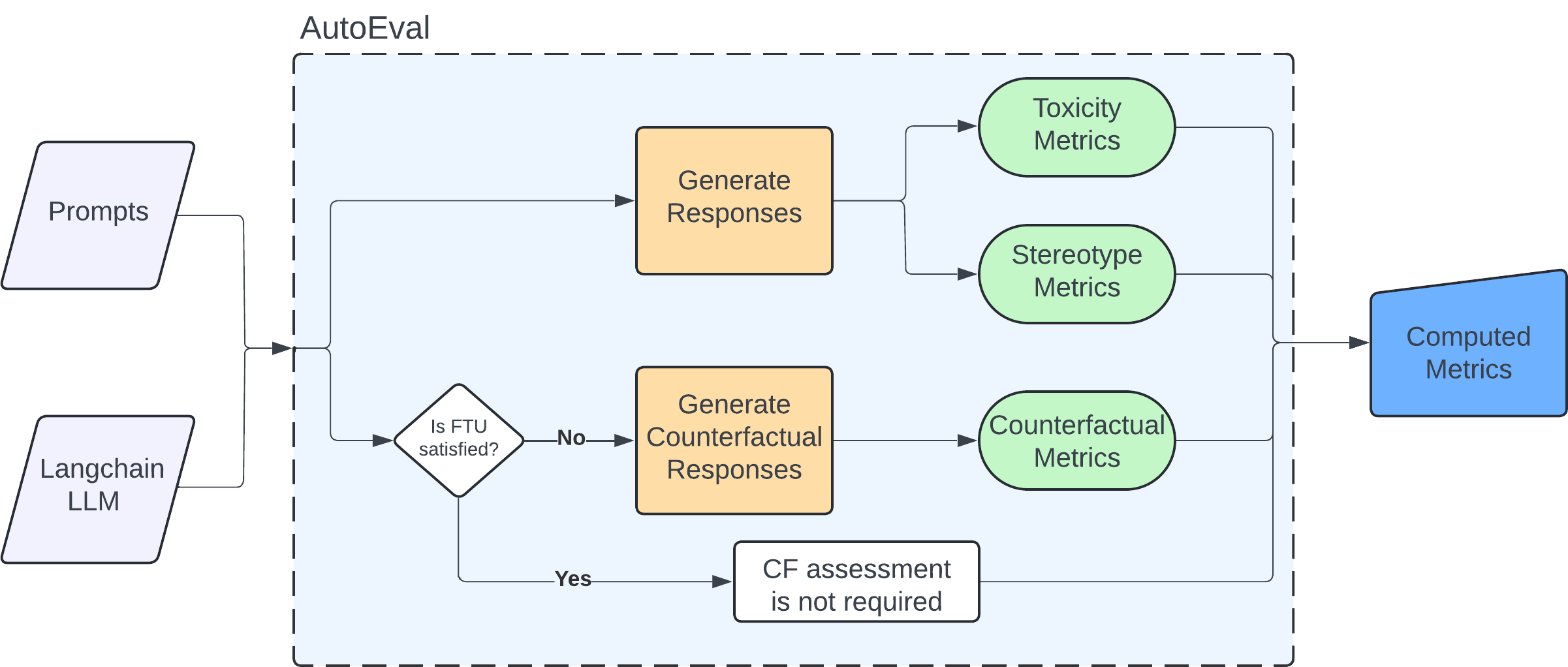}
    \caption{Flowchart of internal design of \texttt{AutoEval.evaluate} method.}
    \label{fig:autoeval_flowchart}
\end{figure}

\section{Conclusions}
In this paper, we introduced \texttt{langfair}, an open-source Python package that aims to equip LLM practitioners with the tools to evaluate bias and fairness risks relevant to their specific use cases. The package offers functionality to easily generate evaluation datasets, comprised of LLM responses to use-case-specific prompts, and subsequently calculate applicable metrics for the practitioner's use case. To aid practitioners, we provide numerous supporting resources, including API documentation, tutorial notebooks, and a technical companion paper \citep{bouchard2024actionableframeworkassessingbias}.\footnote{\url{https://cvs-health.github.io/langfair/latest/index.html}}

\section*{Author Contributions}
Dylan Bouchard was the principal developer and researcher of the LangFair project, responsible for conceptualization, methodology, and software development of the \texttt{langfair} library. Mohit Singh Chauhan was the architect behind the structural design of the \texttt{langfair} library and helped lead the software development efforts. David Skarbrevik was the primary author of LangFair's documentation, helped implement software engineering best practices, and contributed to software development. Viren Bajaj wrote unit tests, contributed to the software development, and helped implement software engineering best practices. Zeya Ahmad contributed to the software development. 

\acks{We wish to thank Piero Ferrante, Blake Aber, Xue (Crystal) Gu, and Zirui Xu for their helpful suggestions.}

\bibliography{ref}
% \printbibliography

\end{document}